\documentclass[10pt,journal,cspaper,compsoc]{IEEEtran}
\ifCLASSOPTIONcompsoc
\else
\fi
\ifCLASSINFOpdf
\else
\fi
\usepackage{times}
\usepackage{epsfig}
\usepackage{graphicx}
\usepackage{amsmath}
\usepackage{amssymb}
\usepackage{rotating}
\usepackage{algorithmic}
\usepackage{algorithm}

\def\a{\mathbf{a}}
\newcommand{\Dt}{\Delta\boldsymbol{\tau}}
\newcommand{\x}{\mathbf{x}}

\newcommand{\g}{\mathbf{g}}
\newcommand{\h}{\mathbf{h}}
\newcommand{\y}{\mathbf{y}}

\newcommand{\s}{\mathbf{s}}

\renewcommand{\P}{\mathbf{P}}
\renewcommand{\l}{\boldsymbol{\zeta}}

\newcommand{\C}{\mathbb{C}}
\renewcommand{\L}{\mathcal{L}}
\renewcommand{\O}{\mathcal{O}}

\renewcommand{\Re}{\mathbb{R}}
\newcommand{\hf}{\hat{\mathbf{h}}}
\newcommand{\gf}{\hat{\mathbf{g}}}

\newcommand{\af}{\hat{\mathbf{a}}}
\def\cf{\hat{\mathbf{s}}}
\newcommand{\xf}{\hat{\mathbf{x}}}
\newcommand{\yf}{\hat{\mathbf{y}}}

\renewcommand{\H}{\mathbf{H}}
\newcommand{\I}{\mathbf{I}}
\newcommand{\conj}{\mbox{conj}}
\newcommand{\lf}{\hat{\boldsymbol{\zeta}}}

\newcommand{\F}{\mathbf{F}}
\newcommand{\A}{\mathbf{A}}

\newcommand{\diag}{\mbox{diag}}

\newcommand{\conv}{\ast}

\begin{document}
%
\title{Correlation Filters with Limited Boundaries}
%
%
%
%

\author{Hamed~Kiani~Galoogahi,
        Terence~Sim,
        and~Simon~Lucey

\textit{\{hkiani, tsim\}@comp.nus.edu.sg, simon.lucey@csiro.au}

}

\IEEEcompsoctitleabstractindextext{%
\begin{abstract}
Correlation filters take advantage of specific properties in
the Fourier domain allowing them to be estimated
efficiently:~$\mathcal{O}(ND \log D)$ in the frequency domain,
versus~$\mathcal{O}(D^{3} + ND^{2})$ spatially where~$D$ is
signal length, and~$N$ is the number of signals.  Recent
extensions to correlation filters, such as MOSSE, have reignited
interest of their use in the vision community due to their robustness
and attractive computational properties. In this
paper we demonstrate, however, that this computational efficiency
comes at a cost. Specifically, we demonstrate that only~$\frac{1}{D}$ proportion of
shifted examples are unaffected by boundary effects which has a dramatic
effect on detection/tracking performance. In this
paper, we propose a novel approach to correlation filter estimation
that: (i) takes advantage of inherent computational redundancies in
the frequency domain, and (ii) dramatically reduces
boundary effects. Impressive object tracking and detection results are
presented in terms of both accuracy and computational efficiency.

\end{abstract}

\begin{keywords}
Correlation filters, object tracking, pattern detection
\end{keywords}}

\maketitle

\IEEEdisplaynotcompsoctitleabstractindextext

%
\IEEEpeerreviewmaketitle

\section{Introduction}
Correlation between two signals is a standard approach to feature
detection/matching. Correlation touches nearly every facet of computer
vision from pattern detection to object tracking. Correlation is
rarely performed naively in the spatial domain. Instead, the fast Fourier
transform (FFT) affords the efficient application of correlating a
desired template/filter with a signal. Contrastingly, however,
most techniques for estimating a template for such a
purpose (i.e. detection/tracking through convolution) are
performed in the spatial domain~\cite{FragTrack,Babenko-2009,Oza,Ross_IJCV2008}.

\begin{figure}
    \begin{center}
     \begin{tabular}{c c}
             \includegraphics[scale=.75]{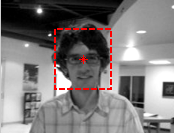} &
             \includegraphics[scale=.75]{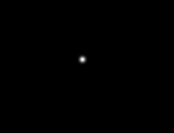}\\
             (a) & (b) \\

             \includegraphics[scale=.85]{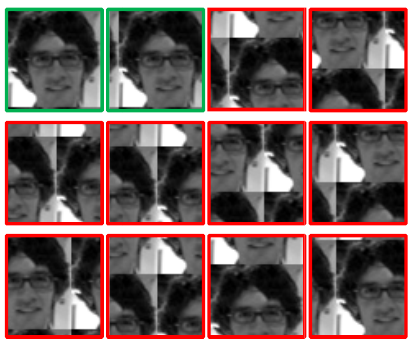} &
             \includegraphics[scale=.85]{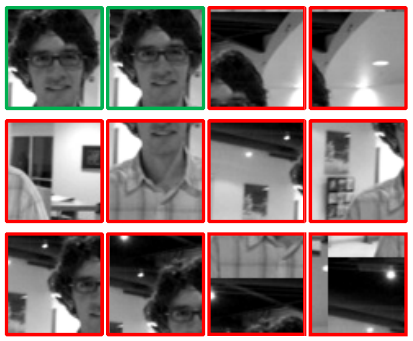}\\
             (c) & (d)
 \end{tabular}
 \end{center}
            \caption{(a) Defines the example of fixed
              spatial support within the image from which the peak
              correlation output should occur. (b) The desired output
              response, based on (a),  of the
              correlation filter when applied to the entire image. (c)
              A subset of patch examples used in a canonical correlation filter
              where green denotes a non-zero correlation output, and
              red denotes a zero correlation output in direct
              accordance with (b). (d) A subset of patch examples used
              in our proposed correlation filter. Note that our
              proposed approach uses patches stemming from different
              parts of the image,
              whereas the canonical correlation filter simply employs
              circular shifted versions of the same single patch. The central
            dilemma in this paper is how to perform (d) efficiently in the
            Fourier domain. The two last patches of (d) show that $ \frac{D-1}{T}$   patches near the image border are affected by circular shift in our method which can be greatly diminished by choosing $D<<T$, where $D$ and $T$ indicate the length of the vectorized face patch in (a) and the image in (a), respectively.}
\label{Fig:Intro}
\end{figure}

This has not always been the case. Correlation filters, developed
initially in the seminal work of Hester and
Casasent~\cite{Hester80}, are a method for learning a
template/filter in the frequency domain that rose to some prominence
in the 80s and 90s. Although many variants have
been proposed~\cite{Hester80,Kumar86,Mahalanobis87,CPR_Book}, the approach's central tenet is to
learn a filter, that when correlated with a set of training signals,
gives a desired response (typically a peak at the origin of the
object, with all other regions of the correlation response map being
suppressed). Like correlation itself, one of the central advantages of
the approach is that it attempts to learn the filter in the frequency
domain due to the efficiency of correlation/convolution in that domain.

Interest in correlation filters has been reignited in the vision world
through the recent work of Bolme et al.~\cite{Bolme10} on Minimum Output Sum of
Squared Error (MOSSE) correlation filters for object detection and
tracking. Bolme et al.'s work was able to circumvent some of the
classical problems with correlation filters and performed well in
tracking under changes in rotation, scale, lighting and partial
occlusion. A central strength of the correlation filter is that it is extremely
efficient in terms of both memory and computation.

\subsection{The Problem}
An unconventional interpretation of a correlation filter, is that of a
discriminative template that has been estimated from an unbalanced set
of ``real-world'' and ``synthetic'' examples. These
synthetic examples are created through the application of a circular
shift on the real-world examples, and are supposed to be representative of those examples at
different translational shifts. We use the term synthetic, as all
these shifted examples are plagued by circular boundary effects and
are not truly representative of the shifted example (see
Figure~\ref{Fig:Intro}(c)). As a result the training set used for learning the template is
extremely unbalanced with one real-world example for every~$D-1$
synthetic examples (where~$D$ is the dimensionality of the examples).

These boundary effects can dramatically affect the resulting performance of the
estimated template. Fortunately, these effects can be largely removed
(see Section~\ref{Sec:CF}) if the correlation filter objective is slightly
augmented, but has to be now solved in the spatial rather than
frequency domains. Unfortunately, this shift to the spatial domain
destroys the computational efficiency that make correlation filters so
attractive. It is this dilemma that is at the heart of our paper.

\subsection{Contribution}
In this paper we make the following contributions:
\begin{itemize}
\item We propose a new correlation filter objective that can drastically reduce the number of
  examples in a correlation filter that are affected by boundary
  effects. We further demonstrate, however, that solving this
  objective in closed form drastically decreases computational
  efficiency:~$\O(D^{3} + ND^{2})$ versus $\O(ND \log D)$ for the canonical
  objective where~$D$ is the length of the vectorized image and~$N$ is the number of
  examples.

\item We demonstrate how this new objective can be
  efficiently optimized in an iterative manner through an Augmented
  Lagrangian Method (ALM) so as to take advantage of inherent
  redundancies in the frequency domain. The efficiency of this new
  approach is~$\O([N+K] T \log T)$ where~$K$ is the number of
  iterations and~$T$ is the size of the search window.

\item We present impressive results for both object detection and
  tracking outperforming MOSSE and other leading non-correlation
  filter methods for object tracking.
\end{itemize}

\subsection{Related Work}
Bolme et al.~\cite{Bolme10} recently proposed an extension to
traditional correlation filters referred to as Minimum Output Sum of
Squared Error (MOSSE) filters. This approach has proven invaluable for
many object tracking tasks, outperforming current state of the art
methods such as~\cite{Babenko-2009, Ross_IJCV2008}. What made the
approach of immediate interest in the vision community was the dramatically faster
frame rates than current state of the art (600 fps versus 30 fps). A strongly related method
to MOSSE was also proposed by Bolme et al.~\cite{Bolme09} for object
detection/localization referred to as Average of Synthetic Exact
Filters (ASEF) which also reported superior performance to state of
the art. A full discussion on other variants of correlation filters such as Optimal Tradeoff
Filters (OTF) ~\cite{Refregier91}, Unconstrained MACE (UMACE) ~\cite{Savvides03} filters, etc. is outside the
scope of this paper. Readers are encouraged to
inspect~\cite{CPR_Book} for a full treatment on the topic.

\subsection{Notation}
Vectors are always presented in lower-case bold (e.g.,~$\a$), Matrices
are in upper-case bold (e.g.,~$\A$) and scalars in
italicized~(e.g. $a$ or~$A$).~$\a(i)$ refers to the~$i$th element of
the vector~$\a$. All~$M$-mode array signals shall be expressed
in vectorized form~$\a$.~$M$-mode arrays are also known as~$M$-mode
matrices, multidimensional matrices, or tensors. We shall be
assuming~$M=2$ mode matrix signals (e.g. $2D$ image arrays) in nearly
all our discussions throughout this paper. This does not preclude,
however, the application of our approach to other~$M \neq 2$ signals.

A~$M$-mode convolution operation is represented as
the $\conv$ operator. One can express a~$M$-dimensional discrete circular
shift~$\Dt$ to a vectorized~$M$-mode matrix $\a$ through the notation
$\a[\Dt]$. The matrix~$\mathbf{I}$ denotes a~$D \times
D$ identity matrix and ~$\mathbf{1}$ denotes a~$D$ dimensional
vector of ones. A~$\hat{}$ applied to any vector denotes the~$M$-mode Discrete
Fourier Transform (DFT) of a vectorized~$M$-mode matrix signal~$\a$ such that~$\af
\leftarrow \mathcal{F}(\a) = \sqrt{D} \F \a$. Where~$\mathcal{F}()$ is the
Fourier transforms operator and $\F$ is the orthonormal~$D \times D$ matrix of
complex basis vectors for mapping to the Fourier domain for any~$D$
dimensional vectorized image/signal. We have chosen to employ a
Fourier representation in this paper due to its particularly useful
ability to represent circular convolutions as a Hadamard product in the Fourier
domain. Additionally, we take advantage of the fact
that~$\diag(\hf)\af = \hf \circ \af$, where $\circ$ represents the
Hadamard product, and~$\diag()$ is an operator that transforms a~$D$
dimensional vector into a~$D \times D$ dimensional diagonal
matrix. The role of filter~$\hf$ or signal~$\af$ can be interchanged
with this property. Any transpose operator~$^{\top}$ on a complex vector
or matrix in this paper additionally takes the complex conjugate in a
similar fashion to the Hermitian adjoint~\cite{CPR_Book}. The
operator~$\conj(\af)$ applies the complex conjugate to the complex vector~$\af$.

\section{Correlation Filters}
\label{Sec:CF}
Due to the efficiency of correlation in the frequency domain,
correlation filters have canonically been posed in the frequency
domain. There is nothing, however, stopping one (other than
computational expense) from expressing a
correlation filter in the spatial domain. In fact, we argue that
viewing a correlation filter in the spatial domain can give: (i)
important links to existing spatial methods for learning
templates/detectors, and (ii) crucial insights into fundamental problems
in current correlation filter methods.

Bolme et. al's~\cite{Bolme10} MOSSE correlation filter can be expressed in the
spatial domain as solving the following ridge regression problem,
\begin{equation}
E(\h) = \frac{1}{2} \sum_{i=1}^{N} \sum_{j =1}^{D} || \y_{i}(j) -
\h^{\top} \x_{i}[\Dt_{j}] ||_{2}^{2}
+ \frac{\lambda}{2} ||\h||_{2}^{2}
\label{Eq:Spatial}
\end{equation}
where~$\y_{i} \in \Re^{D}$ is the desired response for the~$i$-th
observation~$\x_{i} \in \Re^{D}$ and~$\lambda$ is a regularization
term.~$\C = [\Dt_{1},\ldots,\Dt_{D}]$ represents the set of all circular shifts for a
signal of length~$D$. Bolme et al. advocated the use of a 2D Gaussian of small
variance (2-3 pixels) for~$\y_{i}$ centered at the location of the object (typically the
centre of the image patch). The solution to this objective becomes,
\begin{equation}
\h = \H^{-1} \sum_{i=1}^{N} \sum_{j=1}^{D} \y_{i}(j) \x_{i}[\Dt_{j}]
\label{Eq:h_1}
\end{equation}
where,
\begin{equation}
\H =
\lambda \I + \sum_{i=1}^{N} \sum_{j=1}^{D} \x_{i}[\Dt_{j}] \x_{i}[\Dt_{j}]^{\top} \;\;.
\end{equation}
Solving a correlation filter in the spatial domain quickly becomes
intractable as a function of the signal length~$D$, as the
cost of solving Equation~\ref{Eq:h_1} becomes~$\mathcal{O}(D^{3} +
ND^{2})$.

\subsection{Properties}
Putting aside, for now, the issue of computational cost, the
correlation filter objective described in Equation~\ref{Eq:Spatial} produces a
filter that is particularly sensitive to misalignment in
translation. A highly undesirable property when attempting to detect or
track an object in terms of translation. This sensitivity is obtained due to the circular shift
operator~$\x[\Dt]$, where~$\Dt = [\Delta x, \Delta y]^{\top}$ denotes the
2D circular shift in~$x$ and~$y$.

It has been well noted in correlation filter literature~\cite{CPR_Book} that this
circular-shift alone tends to produce filters that do not
generalize well to other types of appearance variation (e.g. illumination, viewpoint,
scale, rotation, etc.). This generalization issue can be somewhat
mitigated through the judicious choice of non-zero regularization
parameter~$\lambda$, and/or through the use of an ensemble~$N > 1$ of
training observations that are representative of the type of
appearance variation one is likely to encounter.

\subsection{Boundary Effects}
A deeper problem with the objective in Equation~\ref{Eq:Spatial},
however, is that the shifted image patches~$\x[\Dt]$ at all values
of~$\Dt \in \C$, except where~$\Dt = \mathbf{0}$, are not
representative of image patches one would encounter in a normal
correlation operation (Figure ~\ref{Fig:Intro}(c)). In signal-processing, one often refers to this
as the~\emph{boundary effect}. One simple way to circumvent this problem spatially is to allow the
training signal~$\x \in \Re^{T}$ to be a larger size than the
filter~$\h \in \Re^{D}$ such that~$T > D$. Through the use of a~$D \times T$
masking matrix~$\P$ one can reformulate Equation~\ref{Eq:Spatial} as,
\begin{equation}
E(\h) = \frac{1}{2}\sum_{i=1}^{N} \sum_{j =1}^{T} || \y_{i}(j) -
\h^{\top} \P \x_{i}[\Dt_{j}] ||_{2}^{2}
+ \frac{\lambda}{2} ||\h||_{2}^{2} \;\;.
\label{Eq:augment}
\end{equation}
The masking matrix~$\P$ of ones and zeros encapsulates what part of
the signal should be active/inactive. The central benefit of this
augmentation in Equation~\ref{Eq:augment} is the dramatic increase in the
proportion of examples unaffected by
boundary effects ($\frac{T - D + 1}{T}$ instead of~$\frac{1}{D}$). From
this insight it becomes clear that if one chooses~$T >> D$ then
boundary effects become greatly diminished (Figure ~\ref{Fig:Intro}(d)). The computational cost~$\mathcal{O}(D^{3} + N T D)$ of
solving this objective is only slightly larger than the cost of
Equation~\ref{Eq:Spatial},
as the role of~$\P$ in practice can be accomplished efficiently through a lookup table.

It is clear in Equation~\ref{Eq:augment}, that boundary effects could be removed completely by
summing over only a~$T-D+1$ subset of all the~$T$ possible circular
shifts. However, as we will see in the following section such a change
along with the introduction of~$\P$ is not possible if we want to
solve this objective efficiently in the frequency domain.

\subsection{Efficiency in the Frequency Domain}
It is well understood in signal processing that
circular convolution in the spatial domain can be expressed as a
Hadamard product in the frequency domain. This allows one to express
the objective in Equation~\ref{Eq:Spatial} more succinctly and
equivalently as,
\begin{eqnarray}
E(\hf) & = & \frac{1}{2} \sum_{i=1}^{N} || \yf_{i} - \xf_{i} \circ \conj(\hf) ||_{2}^{2}
+ \frac{\lambda}{2} ||\hf||_{2}^{2} \label{Eq:frequency}\\
 & = & \frac{1}{2} \sum_{i=1}^{N} || \yf_{i} - \diag(\xf_{i})^{\top} \hf ||_{2}^{2}
+ \frac{\lambda}{2} ||\hf||_{2}^{2} \nonumber \;\;.
\end{eqnarray}
where~$\hf, \xf, \yf$ are the Fourier transforms of~$\h, \x, \y$. The
complex conjugate of~$\hf$ is employed to ensure the operation is
correlation not convolution. The
equivalence between Equations~\ref{Eq:Spatial} and~\ref{Eq:frequency}
also borrows heavily upon another well known property from signal
processing namely, Parseval's theorem which states that
\begin{equation}
\x_{i}^{\top}\x_{j} = D^{-1} \xf_{i}^{\top} \xf_{j} \quad \forall i, j, \quad
\mbox{where } \x \in \Re^{D} \;\;.
\end{equation}
The solution to Equation~\ref{Eq:frequency} becomes
\begin{eqnarray}
\hf & = & [\diag(\cf_{xx}) + \lambda \I]^{-1} \sum_{i=1}^{N} \diag(\xf_{i})\yf_{i} \\
& = & \cf_{xy} \circ^{-1} (\cf_{xx} + \lambda \mathbf{1})
\nonumber
\end{eqnarray}
where $ \circ^{-1}$ denotes element-wise division, and
\begin{equation}
\cf_{xx} = \sum_{i=1}^{N} \xf_{i} \circ \conj(\xf_{i}) \quad \&
\quad \cf_{xy}
= \sum_{i=1}^{N} \yf_{i} \circ \conj(\xf_{i})
\label{Eq:sf}
\end{equation}
are the average auto-spectral and cross-spectral energies respectively of the
training observations. The solution for~$\hf$ in
Equations~\ref{Eq:Spatial} and~\ref{Eq:frequency} are identical (other
than that one is posed in the spatial domain, and the other is in the frequency
domain). The power of this method lies in its computational
efficiency. In the frequency domain a solution to~$\hf$ can be found
with a cost of~$\mathcal{O}(N D \log D)$. The primary cost is associated
with the DFT on the ensemble of training
signals~$\{\x_{i}\}_{i=1}^{N}$ and desired responses
$\{\y_{i}\}_{i=1}^{N}$.

\section{Our Approach}
A problem arises, however, when one attempts to apply the same Fourier
insight to the augmented spatial objective in
Equation~\ref{Eq:augment},
\begin{equation}
E(\h) = \frac{1}{2} \sum_{i=1}^{N} || \yf_{i} - \diag(\xf_{i})^{\top} \sqrt{D}\F\P^{\top}\h ||_{2}^{2}
+ \frac{\lambda}{2} ||\h||_{2}^{2} \;\;.
\label{Eq:naive}
\end{equation}
Unfortunately, since we are enforcing a spatial constraint the
efficiency of this objective balloons to~$\mathcal{O}(D^{3} + N D^{2})$ as~$\h$~\emph{must} be solved in the spatial domain.

\subsection{Augmented Lagrangian}
Our proposed approach for solving Equation~\ref{Eq:naive} involves the
introduction of an auxiliary variable~$\gf$,
\begin{eqnarray}
E(\h,\gf) & = & \frac{1}{2} \sum_{i=1}^{N} || \yf_{i} - \diag(\xf_{i})^{\top} \gf ||_{2}^{2}
+ \frac{\lambda}{2} ||\h||_{2}^{2} \nonumber \\
& \mbox{s.t.} & \gf = \sqrt{D}\F\P^{\top}\h \;\;.
\label{Eq:ecf}
\end{eqnarray}
We propose to handle the introduced equality constraints through an
Augmented Lagrangian Method (ALM)~\cite{boyd_book_2010}. The augmented
Lagrangian of our proposed objective can be formed as,
\begin{eqnarray}
\L(\gf,\h,\lf) & = & \frac{1}{2} \sum_{i=1}^{N} || \yf_{i} -
\diag(\xf_{i})^{\top} \gf ||_{2}^{2} + \frac{\lambda}{2} ||\h||_{2}^{2} \nonumber \\
& & + \quad \lf^{\top}(\gf - \sqrt{D} \F \P^{\top} \h ) \nonumber \\
& & + \quad \frac{\mu}{2} ||\gf - \sqrt{D} \F \P^{\top} \h ||_{2}^{2}
\end{eqnarray}
where~$\mu$ is the penalty factor that controls the rate of convergence of the ALM, and~$\lf$ is
the Fourier transform of the Lagrangian vector needed
to enforce the newly introduced equality constraint in
Equation~\ref{Eq:ecf}. ALMs are not new
to learning and computer vision, and have recently been used to great
effect in a number of
applications~\cite{boyd_book_2010,delbue_PAMI_2011}. Specifically, the
Alternating Direction Method of Multipliers (ADMMs) has provided a
simple but powerful algorithm that is well suited to distributed
convex optimization for large learning and vision problems. A full
description of ADMMs is outside the scope of this paper (readers are
encouraged to inspect~\cite{boyd_book_2010} for a full treatment and
review), but they can be loosely interpreted as applying a
Gauss-Seidel optimization strategy to the augmented Lagrangian
objective. Such a strategy is advantageous as it often leads to
extremely efficient subproblem decompositions. A full description of our proposed algorithm can be seen in
Algorithm~\ref{Alg:Ours}. We detail each of the subproblems as follows:

\subsection{Subproblem g}
\begin{eqnarray}
\gf^{*} & = & \arg \min \L(\gf; \hf, \lf) \label{Eq:g} \\
          & = & (\cf_{xy} + \mu \hf - \lf)\circ^{-1} (\cf_{xx} + \mu\mathbf{1}) \nonumber
\end{eqnarray}
where~$\hf = \sqrt{D}\F\P^{\top}\h$. In practice~$\hf$ can be estimated extremely
efficiently by applying a FFT to~$\h$ padded with zeros implied by
the~$\P^{\top}$ masking matrix.

\subsection{Subproblem h}
\begin{eqnarray}
\h^{*} & = & \arg \min \L(\h; \g, \l) \label{Eq:h} \\
          & = & (\mu + \frac{\lambda}{\sqrt{D}})^{-1}(\mu \g + \l)  \nonumber
\end{eqnarray}
where~$\g = \frac{1}{\sqrt{D}}\P\F^{\top}\gf$ and $\l =
\frac{1}{\sqrt{D}}\P\F^{\top}\lf$. In practice both~$\g$ and~$\l$ can be
estimated extremely efficiently by applying an inverse FFT and then
applying the lookup table implied by the masking matrix~$\P$.

\subsection{Lagrange Multiplier Update}
\begin{equation}
\lf^{(i+1)} \leftarrow \lf^{(i)} + \mu(\gf^{(i+1)} - \hf^{(i+1)})
\label{Eq:l}
\end{equation}
where~$\gf^{(i+1)}$ and~$\hf^{(i+1)}$ are the current solutions to the
above subproblems at iteration~$i+1$ within the iterative ADMM.

\subsection{Choice of~$\boldsymbol{\mu}$}
A simple and common~\cite{boyd_book_2010} scheme for selecting~$\mu$ is the following
\begin{equation}
\mu^{(i+1)} = \min(\mu_{\max},\beta \mu^{(i)}) \;\;.
\label{Eq:mu}
\end{equation}
We found experimentally~$\mu^{(0)} = 10^{-2}$, $\beta = 1.1$
and~$\mu_{\max} = 20$ to perform well.

\subsection{Computational Cost}
Inspecting Algorithm~\ref{Alg:Ours} the dominant cost per iteration of
the ADMM optimization process is~$\O(T \log T)$ for FFT. There is a
pre-computation cost (before the iterative component, steps 4 and 5) in the algorithm for estimating the auto- and
cross-spectral energy vectors~$\cf_{xx}$ and~$\cf_{xy}$
respectively. This cost is~$\O(NT \log T)$ where~$N$ refers to the
number of training signals. Given that~$K$ represents the number of
ADMM iterations the overall cost of the algorithm is
therefore~$\O([N + K]T\log T)$.

\begin{algorithm}
\caption{Our approach using ADMMs}
\begin{algorithmic}[1]
  \vspace{1mm}
  \STATE Intialize $\h^{(0)}$, $\l^{(0)}$.
  \STATE Pad with zeros and apply FFT:~$\sqrt{D}\F\P^{\top} \h^{(0)} \rightarrow
  \hf^{(0)}$.
  \STATE Apply FFT:~$\sqrt{D}\F\l^{(0)} \rightarrow \lf^{(0)}$.
   \STATE Estimate auto-spectral energy $\cf_{xx}$ using Eqn. (\ref{Eq:sf}).
   \STATE Estimate cross-spectral energy $\cf_{xy}$ using Eqn. (\ref{Eq:sf}).
  \STATE $i = 0$
  \REPEAT
 \STATE Solve for $\gf^{(i+1)}$ using Eqn.  (\ref{Eq:g}), $\hf^{(i)}$ \& $\lf^{(i)}$.
 \STATE Inverse FFT then crop:~$\frac{1}{\sqrt{D}} \P \F^{\top}
 \gf^{(i+1)} \rightarrow \g^{(i+1)}$.
 \STATE Inverse FFT then crop:~$\frac{1}{\sqrt{D}} \P \F^{\top}
 \lf^{(i+1)} \rightarrow \l^{(i+1)}$.
 \STATE Solve for $\h^{(i+1)}$ using Eqn.  (\ref{Eq:h}), $\g^{(i+1)}$ \& $\l^{(i)}$.
  \STATE Pad and apply FFT:~$\sqrt{D}\F\P^{\top} \h^{(i+1)} \rightarrow
  \hf^{(i+1)}$.
  \STATE Update Lagrange multiplier vector Eqn. (\ref{Eq:l}).
  \STATE Update penalty factor Eqn. (\ref{Eq:mu}).
  \STATE $i = i + 1$
 \UNTIL {$\gf, \h, \lf$ has converged}
 \vspace{1mm}
\end{algorithmic}
\label{Alg:Ours}
\end{algorithm}

\section{Experiments}

\subsection{Localization Performance}
In the first experiment, we evaluated our method on the problem of eye localization,
comparing with prior correlation filters, e.g. OTF ~\cite{Refregier91},
MACE ~\cite{Mahalanobis87}, UMACE ~\cite{Savvides03}, ASEF ~\cite{Bolme09}, and MOSSE ~\cite{Bolme10}.
 The CMU Multi-PIE face database \footnote{http://www.multipie.org/} was used for this experiment, containing 900 frontal faces with neutral expression and normal illumination. We randomly selected 400 of these images for training and the reminder for testing. All images were cropped to have a same size of $ 128 \times 128 $ such that the left and right eye are respectively centered at (40,32) and (40,96) coordinates. The cropped images were power normalized to have a zero-mean and standard deviation of 1.0. Then, a 2D cosine window was employed to reduce the frequency effects caused by opposite borders of the images in the Fourier domain.

We trained a $ 64 \times 64 $ filter of the right eye using $ 64 \times 64 $ cropped patches (centered upon the right eye) for the other methods, and full face images for our method ($T=128 \times 128$ and $D=64 \times 64$). Similar to ASEF and MOSSE, we defined the desired response as a 2D Gaussian function with an spatial variance of $ s = 2$. Eye localization was
performed by correlating the filters over the testing images followed by selecting the peak of the output as the predicted eye location. The eye localization was evaluated by the distance between the predicted and desired eye locations normalized by inter-ocular distance,
\begin{equation}
d = \frac{\Vert \mathbf{p}_{r} - \mathbf{m}_{r} \Vert_{2}}{\Vert \mathbf{m}_l - \mathbf{m}_r \Vert_{2}}
\end{equation}
where $ \mathbf{m}_r $ and $ \mathbf{m}_l $ respectively indicate the true coordinates of the right and left eye, and $ \mathbf{p}_r $ is the predicted location of the right eye. A localization with normalized distance $ d < \emph{th} $ was considered as a successful localization. The threshold \emph{th} was set to a fraction of inter-ocular distance.

\begin{figure}
    \begin{center}
         \begin{tabular}{@{}c@{} @{}c@{} }
          \multicolumn{2}{c}{\includegraphics[scale=.65]{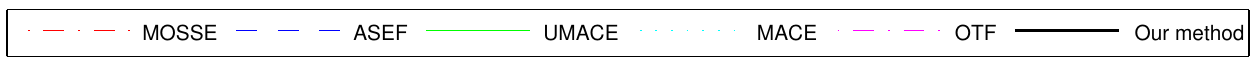}}\\

            \includegraphics[scale=.65]{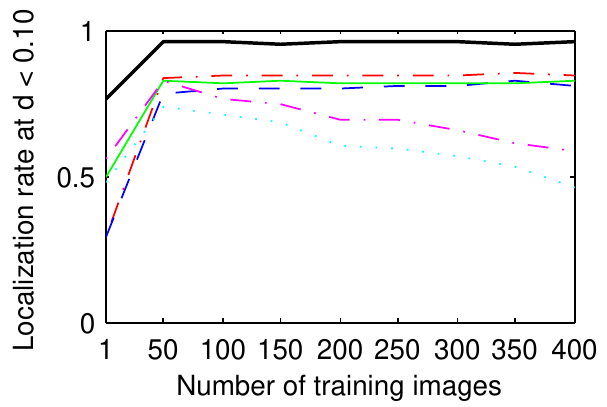} &
            \includegraphics[scale=.65]{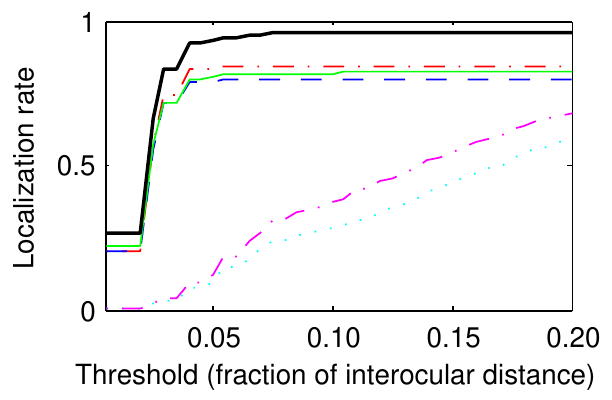} \\
            (a) & (b)
             \end{tabular}
     \end{center}
            \caption{ Eye localization performance as a function of (a) number of training images, and (b) localization thresholds. }
\label{Fig:eye_loc_acc}
\end{figure}

\begin{figure}
\begin{center}
\begin{tabular}{ c@{\hskip 1mm} c @{\hskip 1mm}c @{\hskip 1mm}c@{\hskip 1mm} c@{\hskip 1mm} }

     ~ &   \footnotesize UMACE &  \footnotesize ASEF &   \footnotesize MOSSE &   \footnotesize Our method \\
    ~ &
      \includegraphics[scale=.25]{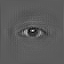} &
      \includegraphics[scale=.25]{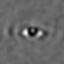} &
      \includegraphics[scale=.25]{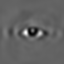} &
      \includegraphics[scale=.25]{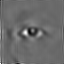} \\

    \includegraphics[scale=.4]{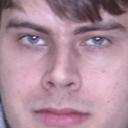} &
       \includegraphics[scale=.4]{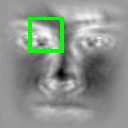} &
      \includegraphics[scale=.4]{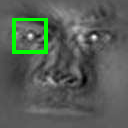} &
      \includegraphics[scale=.4]{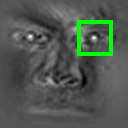} &
       \includegraphics[scale=.4]{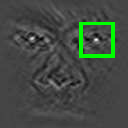} \\

   ~ &   \footnotesize PSR = 3.1 &  \footnotesize PSR = 8.4 &   \footnotesize PSR = 9.3 &   \footnotesize PSR = 15.7

    \end{tabular}

    \end{center}
    \caption{ An example of eye localization is shown for an image with normal lighting. The outputs (bottom) are
  produced using 64x64 correlation filters (top). The green box represents the approximated location of the right eye (output peak). The peak strength measured by PSR shows the sharpness of the output peak.}
    \label{Fig:visloc}
    \end{figure}

 The average of evaluation results across 10 random runs are depicted in Figure~\ref{Fig:eye_loc_acc},
 where our method outperforms the other approaches across all thresholds and training set sizes. The accuracy of OTF and MACE declines by increasing the number of training images
 due to over-fitting. During the experiment, we observed that the low performance of
 the UMACE, ASEF and MOSSE was mainly caused by wrong localizations of the left eye and the nose.
 This was not the case for our method, as our filter was trained in a way that return zero correlation
 values when centred upon non-target patches of the face image. A visual depiction of the filters and their outputs can be seen in Figure~\ref{Fig:visloc},
 illustrating examples of wrong and correct localizations.
 The Peak-to-Sidelobe Ratio (PSR) ~\cite{Bolme10} values show
 that our method returns stronger output compared to the other filters.

\begin{figure}
    \begin{center}
         \begin{tabular}{@{}c  @{} c@{}}
            \includegraphics[scale=.6]{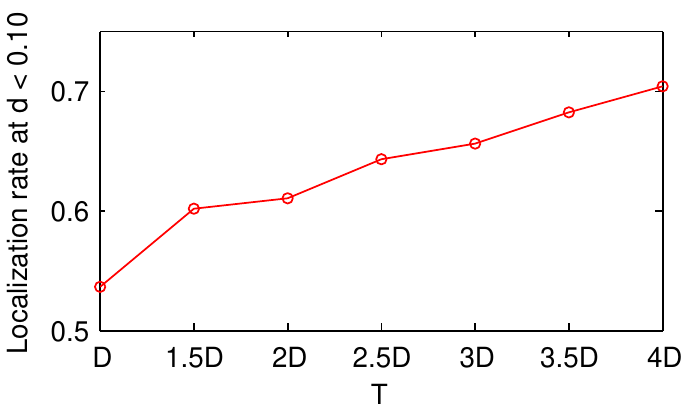} &
             \includegraphics[scale=.62]{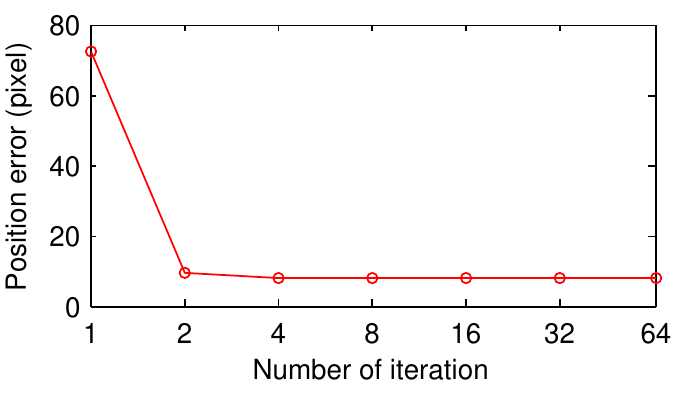} \\
             (a) & (b)

              \end{tabular}
     \end{center}
            \caption{(a) The localization rate obtained by different sizes of training images (\emph{T}), the size of the trained filter is $D = 32 \times 32.$ (b) The position error of tracking versus the number of ADMM iterations. We selected 4 iterations as  a tradeoff between tracking performance and computation.}
\label{Fig:TvsD}
\end{figure}

 Moreover, we examined the influence of \emph{T} (the size of training images) on the performance of eye localization. For this purpose, we employed cropped patches of the right eye with varying sizes of $ T = \{D, 1.5D, 2D, 2.5D, 3D, 3.5D, 4D\}$ to train filters of size $ D = 32 \times 32$. The localization results are illustrated in Figure ~\ref{Fig:TvsD}(a), showing that the lowest performance obtained when \emph{T }is equal to \emph{D} and the localization rate improved by increasing the size of the training patches with respect to the filter size. The reason is that by choosing $T > D $ the portion of patches unaffected by boundary effects $ (\frac{T-D+1}{T})$ reduces.

\subsection{Runtime Performance}
\begin{figure}
    \begin{center}
     \begin{tabular}{@{}c@{} c@{}}
             \includegraphics[scale=.67]{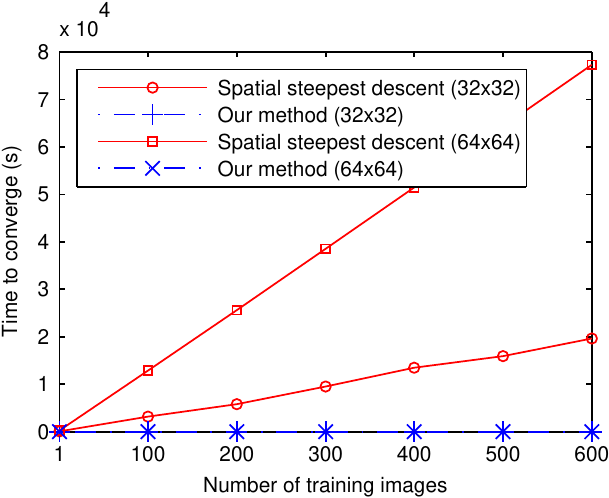} &
             \includegraphics[scale=.67]{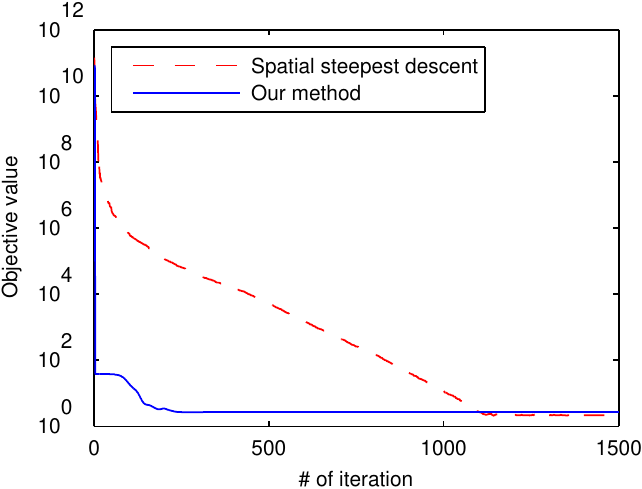}\\
             (a) & (b)
 \end{tabular}
 \end{center}
            \caption{Runtime performance of our method
              against another naive iterative method (steepest descent
              method)~\cite{zeiler_CVPR_2010}. Our approach
              enjoys superior performance in terms of: (a) convergence
              speed to train two filters with different sizes (32x32 and 64x64) and (b) the number of iterations required to converge.}
\label{fig:zeiler_objval}
\end{figure}

This experiment demonstrates the advantage of our approach to other iterative methods. Specifically, we compared our proposed approach against other methods in literature for learning filters efficiently using iterative methods. We compared our
convergence performance with a steepest descent method~\cite{zeiler_CVPR_2010} for optimizing our same
objective. Results can be seen in Figure \ref{fig:zeiler_objval}: (a)
represents time to converge as a function of the number of training
images, and (b) represents the number of iterations required to
optimize the objective (in Equation~\ref{Eq:naive}).

In (a) one notices impressively how convergence performance is largely
independent to the number of images used during training. This can
largely be attributed to the pre-computation of the auto- and
cross-spectral energy vectors. As a result, iterations of the ADMM do not need to
re-touch the training set, allowing our proposed approach to
dramatically outperform more naive iterative approaches. Similarly, in (b) one also
notices how relatively few iterations are required to achieve good
convergence.

\subsection{Tracking Performance}

Finally, we evaluated the proposed method for the task of real-time
tracking on a sequence of commonly used test videos~\cite{Ross_IJCV2008}, described in~Table~\ref{videos}. We compared our approach with state-of-the-art trackers including MOSSE ~\cite{Bolme10},
kernel-MOSSE ~\cite{Henriques12}, MILTrack ~\cite{Babenko2011}, STRUCK ~\cite{Hare2011},
OAB ~\cite{Grabner2006}, SemiBoost ~\cite{Grabner2008}, FragTrack ~\cite{FragTrack} and
IVT ~\cite{Ross_IJCV2008}. All of these methods were tuned by the parameter settings proposed
in their reference papers. The desired response for a $m \times n$ target was defined as a 2D Gaussian with a
variance of $ \s = \sqrt{mn}/16 $. The regularization parameter $ \lambda $ was set to $ 10^{-2} $.
We evaluated our method with differen number of iterations $ \{1, 2, 4, 8, 16, 32, 64\}$, as shown in Figure~\ref{Fig:TvsD}(b), and eventually selected four iterations (a tradeoff between precision and tracking speed) for our tracker.
A track initialization process was employed for our approach and MOSSE, where eight random affine
perturbations were used to initialize the first filter. We borrowed the online adaption from the work
of Bolme et al. ~\cite{Bolme10} to adapt our filter at $i^{th}$ frame using averaged auto-spectral and cross-spectral energies:

\begin{eqnarray}
(\cf_{xx})^{i} &=& \eta (\xf_{i} \circ \conj(\xf_{i})) + (1- \eta) (\cf_{xx})^{i-1} \nonumber \\\
(\cf_{xy})^{i} &=& \eta (\yf_{i} \circ \conj(\xf_{i})) + (1-\eta)(\cf_{xy})^{i-1}
\label{Eq:adaption}
\end{eqnarray}
where,  $ \eta$ is the adaption rate. We practically found that $\eta = 0.025$ is appropriate for our method to quickly be adapted against pose change, scale, illumination, etc.

The tracking results are evaluated in Table ~\ref{Table:tracking_res} following the recent tracking papers \cite{Babenko2011}~\cite{Hare2011}~\cite{Grabner2006}, including (\romannumeral 1)
percentage of frames where the predicted position is within 20 pixels of the ground truth
(precision), (\romannumeral 2) average localization error in pixels, and (\romannumeral 3 )
tracking speed as frames per second (\emph{fps}) Our method averagely achieved maximum precisions and
minimum localization errors, followed by STRUCK. One explanation for this is that our method
employs a rich set of training samples containing all possible positive (target) and negative
(non-target) patches to train the correlation filter. Whilst, the non filter approaches such as
STRUCK and MILTrack are limited by learning a small subset of positive and negative patches.
Similarly, it can be explained that the accuracy of MOSSE and kernel-MOSSE are affected
by using synthetic negative samples which are not representative of the "real-world" examples,
as illustrated in Figure~\ref{Fig:Intro}(c). Moreover, our method owes its robustness against
challenging variations in scale (\emph{Cliffbar} and \emph{Twinings}), illumination (\emph{Sylv}),
pose (\emph{David}), appearance (\emph{Girl}) and partial occlusion (\emph{Faceocc1} and \emph{Faceocc2})
to the online adaption.
In the case of tracking speed, MOSSE outperformed the other methods by 600 \emph{fps}.
Our method obtained lower \emph{fps} compared to MOSSE and kernel-MOSSE, due to its iterative manner.
However, it obtained a tracking speed of 50 \emph{fps} which is appropriate for real-time tracking.

\begin{figure*}
    \begin{center}
    \begin{tabular}{@{}c@{}  @{}c@{} @{}c@{} @{}c@{}}
    \setlength{\tabcolsep}{0pt}
    Coke & Clifbar & David & Faceocc2 \\
             \includegraphics[scale=.65]{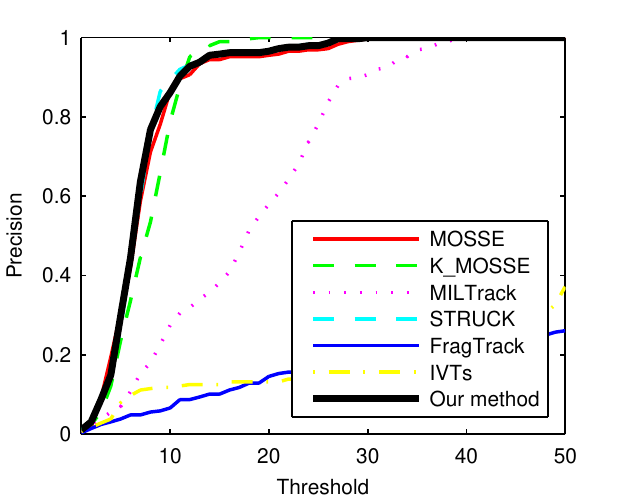} &
             \includegraphics[scale=.65]{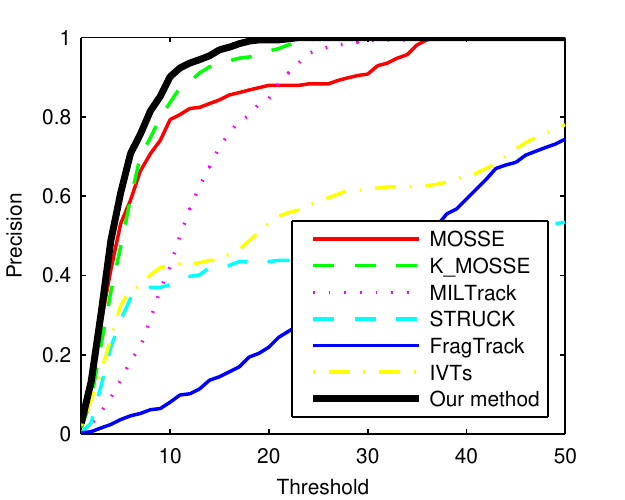}  &
             \includegraphics[scale=.65]{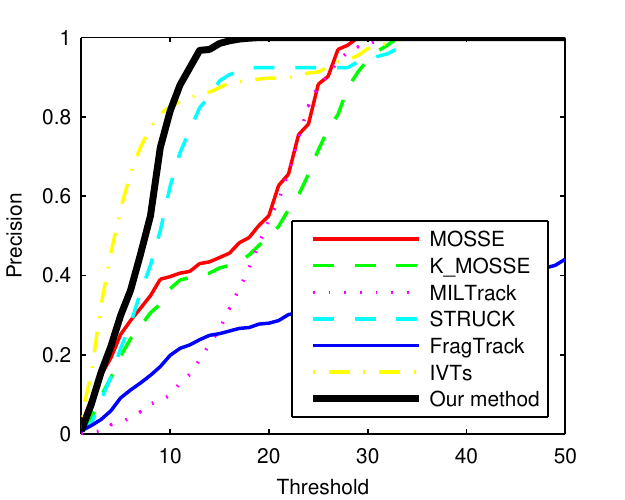}  &
             \includegraphics[scale=.65]{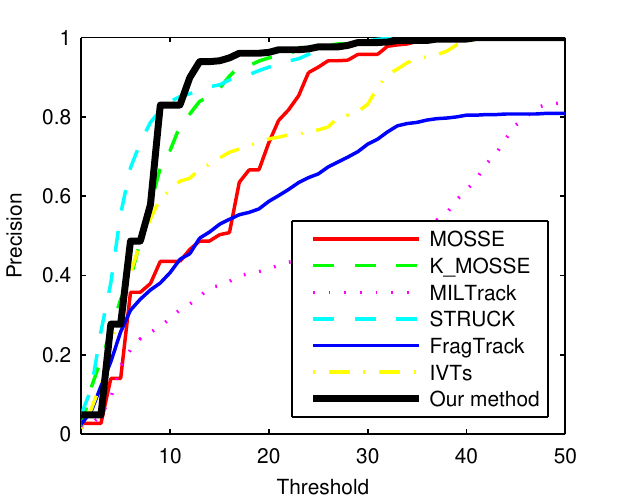}  \\
   Girl & Sylv & Twinings & Tiger1 \\
            \includegraphics[scale=.65]{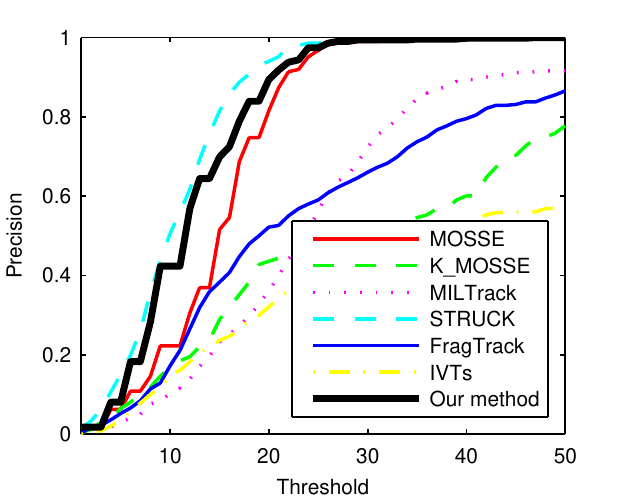} &
            \includegraphics[scale=.65]{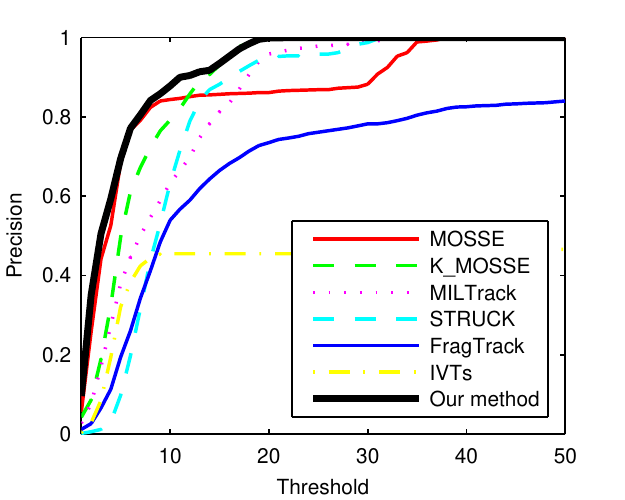}  &
            \includegraphics[scale=.65]{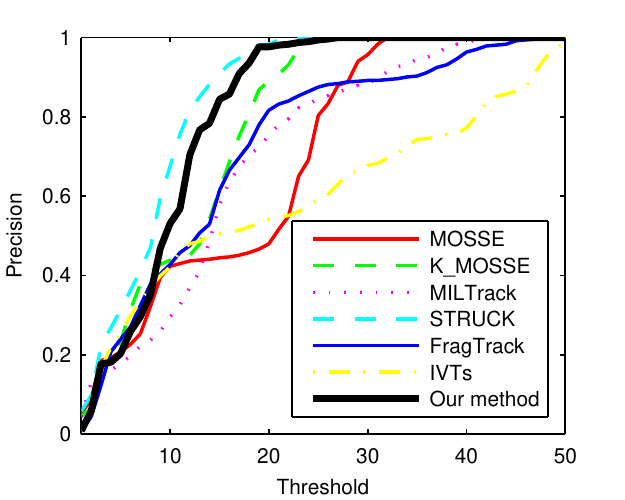} &
            \includegraphics[scale=.65]{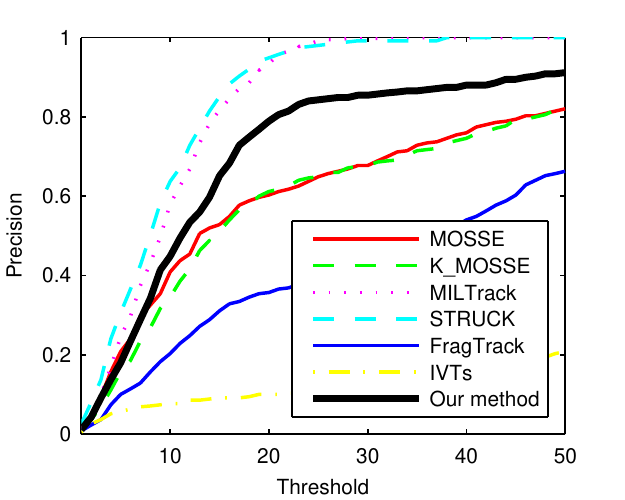}  \\
         \multicolumn{4}{c}{(a)} \\

    Coke & Clifbar & David & Faceocc2 \\
             \includegraphics[scale=.65]{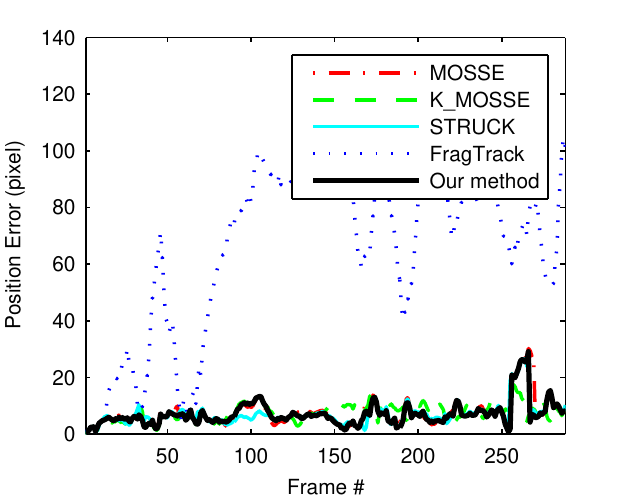} &
             \includegraphics[scale=.65]{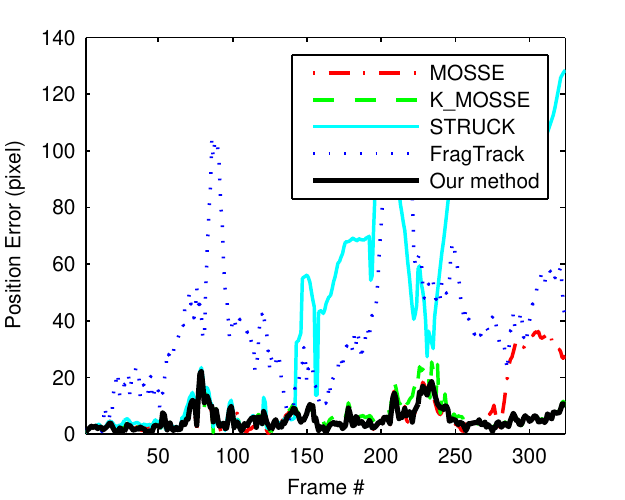} &
             \includegraphics[scale=.65]{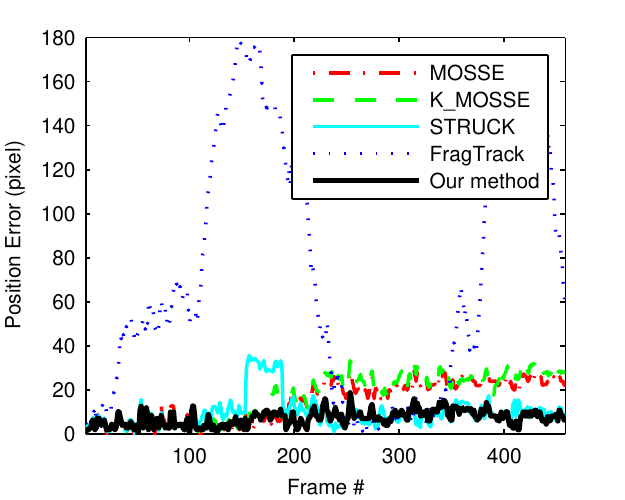} &
             \includegraphics[scale=.65]{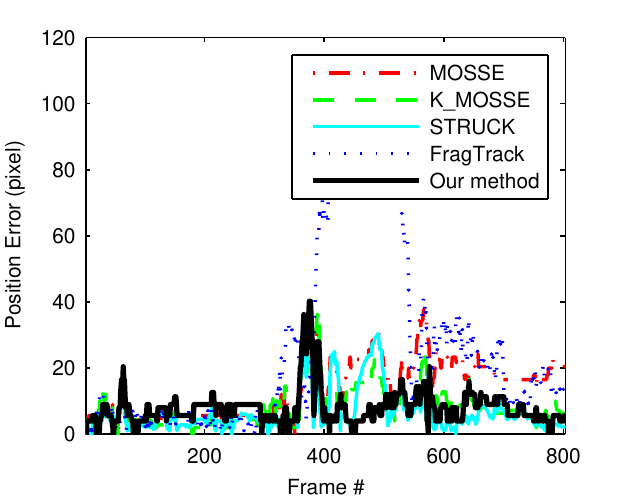} \\

     Girl & Sylv & Twinings & Tiger1 \\
             \includegraphics[scale=.65]{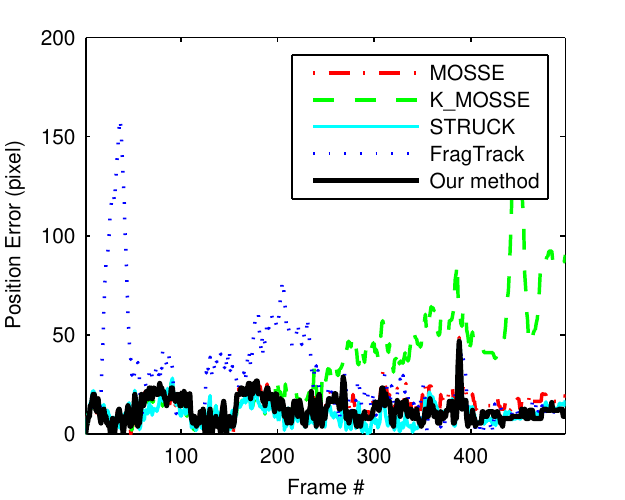} &
             \includegraphics[scale=.65]{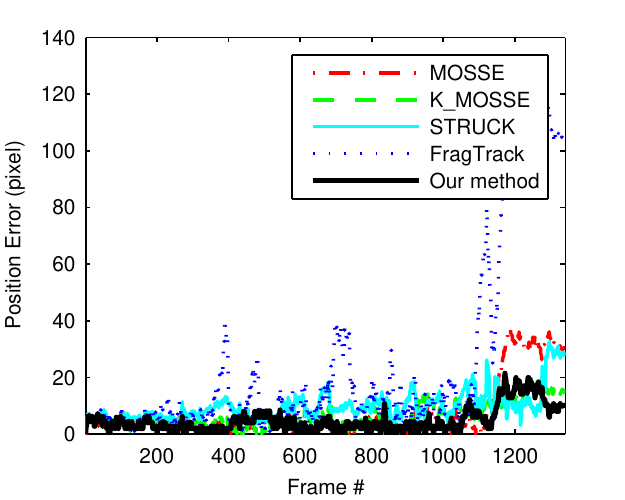} &
             \includegraphics[scale=.65]{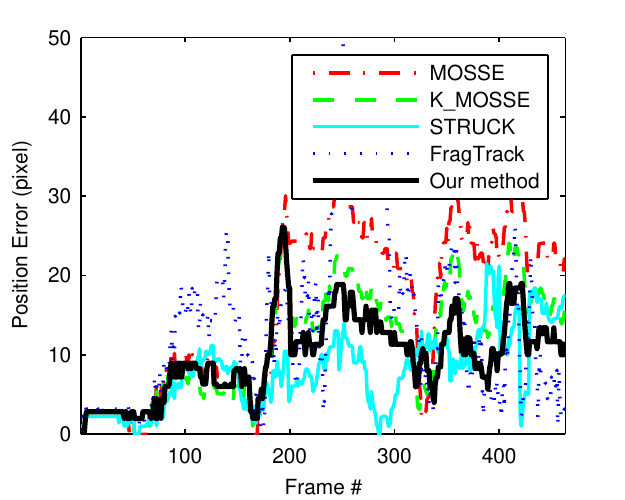} &
             \includegraphics[scale=.65]{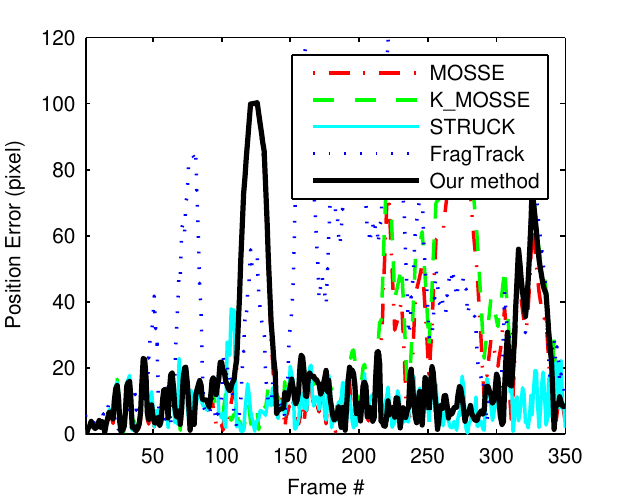} \\
        \multicolumn{4}{c}{(b)}

      \end{tabular}
    \end{center}
           \caption{Tracking results for selected videos, (a) precision versus the thresholds, and (b) position error per frame.}
           \label{Fig:track}
\end{figure*}

\begin{figure*}
    \begin{center}
    \begin{tabular}{@{\hskip 1mm}c@{\hskip 1mm} @{\hskip 1mm}c@{\hskip 1mm} @{\hskip 1mm}c@{\hskip 1mm} @{\hskip 1mm}c@{\hskip 1mm} @{\hskip 1mm}c@{\hskip 1mm} }

            \multicolumn{5}{c}{Cliffbar} \\

            \includegraphics[scale=.45]{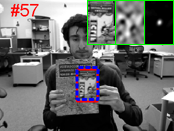} &
             \includegraphics[scale=.45]{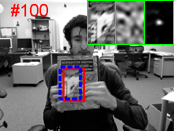}  &
             \includegraphics[scale=.45]{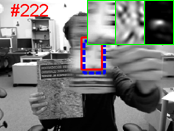} &
             \includegraphics[scale=.45]{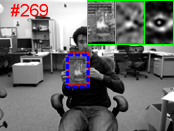} &
             \includegraphics[scale=.45]{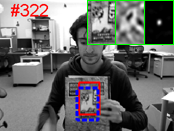}  \\

%

            \multicolumn{5}{c}{David} \\
            \includegraphics[scale=.45]{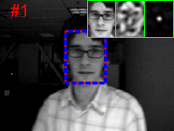} &
            \includegraphics[scale=.45]{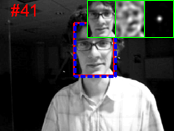} &
            \includegraphics[scale=.45]{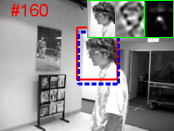}  &
           \includegraphics[scale=.45]{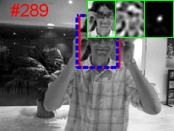} &
             \includegraphics[scale=.45]{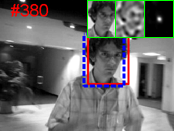} \\

           \multicolumn{5}{c}{Faceocc2} \\
             \includegraphics[scale=.45]{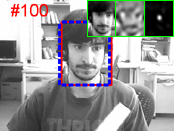} &
             \includegraphics[scale=.45]{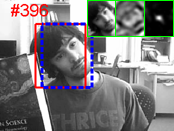}  &
             \includegraphics[scale=.45]{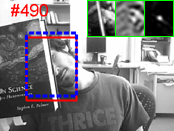} &
             \includegraphics[scale=.45]{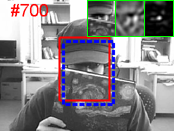} &
            \includegraphics[scale=.45]{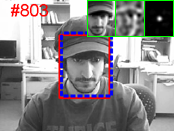} \\

          \multicolumn{5}{c}{Girl} \\
             \includegraphics[scale=.45]{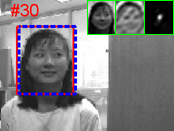} &
             \includegraphics[scale=.45]{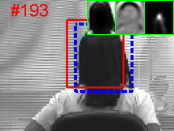}  &
             \includegraphics[scale=.45]{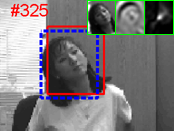} &
              \includegraphics[scale=.45]{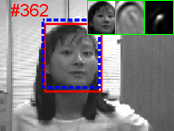} &
             \includegraphics[scale=.45]{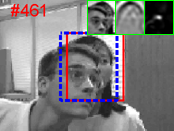} \\
          \multicolumn{5}{c}{Sylv} \\
            \includegraphics[scale=.45]{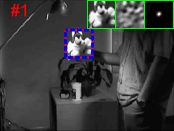} &
             \includegraphics[scale=.45]{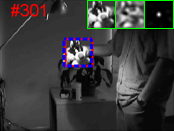} &
             \includegraphics[scale=.45]{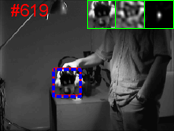}  &
             \includegraphics[scale=.45]{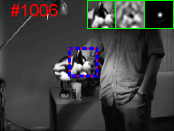} &
             \includegraphics[scale=.45]{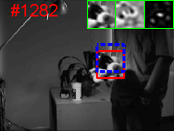} \\
%

      \end{tabular}
    \end{center}
           \caption{Tracking results of our method over the test videos with challenging variations of pose, scale, illumination and partial occlusion. The blue (dashed) and red boxes respectively represent the ground truth and the positions predicted by our method. For each frame, we illustrate the target, trained filter and correlation output.}
           \label{Fig:track1}
\end{figure*}

A visual depiction of tracking results for some selected videos is shown in
Figures~\ref{Fig:track} and ~\ref{Fig:track1}, where our method achieved higher precisions over all videos except \emph{Tiger1} and \emph{Twinings}. Moreover, Figure~\ref{Fig:track}(b) shows that our approach suffers from less drift over the selected test videos.

\begin{table}
\begin{center}

    \begin{tabular}{|r|c|l|}
        \hline
        \textbf{Sequence} & \textbf{Frames} & \textbf{Main Challenges} \\ \hline \hline
        \textbf{Faceocc1} &	886	& Moving camera, occlusion \\ \hline
       \textbf{Faceocc2} & 812	& Appearance change, occlusion \\ \hline
        \textbf{Girl}	& 502	&  Moving camera, scale change \\ \hline
       \textbf{Sylv}	& 1344 &	Illumination and pose change \\ \hline
       \textbf{Tiger1}	& 354	& Fast motion, pose change \\ \hline
       \textbf{David} & 462 &	Moving camera, illumination change \\ \hline
        \textbf{Cliffbar}	& 472 &	Scale change,  motion blur \\ \hline
        \textbf{Coke Can}	& 292 &	Illumination change, occlusion \\ \hline
       \textbf{Dollar} &	327	& Similar object, appearance change \\ \hline
        \textbf{Twinings}	& 472 &	Scale and pose change \\
        \hline
    \end{tabular}
\end{center}
\caption {Video sequences used for tracking evaluation.}
\label{videos}
\end{table}

\begin{table*}
\begin{center}
    \begin{tabular}{|r|c|c|c|c|c|c|c|c|}
        \hline
        ~ & \textbf{MOSSE} &	\textbf{KMOSSE} & \textbf{MILTrack} & \textbf{STRUCK} & \textbf{OAB(1)} & \textbf{SemiBoost} & \textbf{FragTrack} & \textbf{Our method} \\ \hline

        \textbf{FaceOcc1} & \{\textbf{1.00}, 7\} & \{\textbf{1.00}, \textbf{5}\} & \{0.75, 17\} & \{0.97, 8\} & \{0.22, 43\} & \{0.97, 7\} & \{0.94, 7\} &  \{\textbf{1.00}, 8\} \\ \hline

        \textbf{FaceOcc2} & \{0.74, 13\} & \{0.95, 8\} & \{0.42, 31\} & \{0.93, \textbf{7}\} & \{0.61, 21\} & \{0.60, 23\} & \{0.59, 27\}  & \{\textbf{ 0.97}, \textbf{7}\} \\ \hline

        \textbf{Girl} & \{0.82, 14\} & \{0.44, 35\} & \{0.37, 29\} & \{\textbf{0.94}, \textbf{10}\} & - & - & \{0.53, 27\} & \{0.90, 12\} \\ \hline

        \textbf{Sylv} & \{0.87, 7\} & \{\textbf{1.00}, 6\} & \{0.96, 8\} & \{0.95, 9\} & \{0.64, 25\} & \{0.69, 16\} & \{0.74, 25\}   & \{\textbf{1.00}, \textbf{4}\} \\ \hline

        \textbf{Tiger1} & \{0.61, 25\} & \{0.62, 25\} & \{0.94, 9\} & \{\textbf{0.95}, \textbf{9}\} & \{0.48, 35\} & \{0.44, 42\} & \{0.36, 39\}   & \{0.79, 18\} \\ \hline

        \textbf{David}	& \{0.56, 14\} & \{0.50, 16\}	& \{0.54, 18\} & \{0.93, 9\} & \{0.16, 49\} & \{0.46, 39\} & \{0.28, 72\} & \{\textbf{1.00}, \textbf{7}\} \\ \hline

        \textbf{Cliffbar} &	\{0.88, 8\} & \{0.97, 6\} & \{0.85, 12\} & \{0.44, 46\} &	\{0.76, -\} &	- & \{0.22, 39\}& \{\textbf{ 1.00}, \textbf{5}\} \\ \hline

        \textbf{Coke Can} &	\{0.96, \textbf{7}\}	& \{\textbf{1.00}, \textbf{7}\} & \{0.58, 17\} & \{0.97, \textbf{7}\} & \{0.45, 25\} & \{0.78, 13\} & \{0.15, 66\} &   \{0.97, \textbf{7}\} \\ \hline

        \textbf{Dollar} & \{\textbf{1.00}, \textbf{4}\} & \{\textbf{1.00}, \textbf{4}\}& \{\textbf{1.00}, 7\} & \{\textbf{1.00}, 13\} & \{0.67, 25\} & \{0.37, 67\} & \{0.40, 55\} & \{\textbf{ 1.00}, 6\} \\ \hline

        \textbf{Twinings} & \{0.48, 16\} & \{0.89, 11\} & \{0.76, 15\} & \{\textbf{0.99}, \textbf{7}\} & \{0.74, -\} & - & \{0.82, 14\} &  \{\textbf{0.99}, 9\} \\ \hline \hline

        \textbf{\textit{mean}} &	\{0.80, 11\}	& \{0.84, 12\} & \{0.72, 16\} & \{0.91, 12\} & \{0.53, 31\} & \{0.62, 29\} & \{0.51, 37\}   & \{\textbf{0.97}, \textbf{8}\} \\ \hline \hline
         \textbf{\textit{fps}} &	\textbf{600}	& 100 & 25 & 11 & 25 & 25 & 2  & 50 \\  \hline
    \end{tabular}
\end{center}
\caption {The tracking performance is shown as a tuple of \{\emph{precision within 20 pixels},
\emph{average position error in pixels}\}, where our method achieved the best performance over 8 of 10 videos.
The best \emph{fps} was obtained by MOSSE. Our method obtained a real-time tacking speed
of 50 \emph{fps} using four iterations of ADMM. The best result for each video is highlighted in bold. }
\label{Table:tracking_res}
\end{table*}

\hspace{-10mm}

\vspace{-3mm}

\vspace{-3mm}

\section{Conclusions}
A method for estimating a correlation filter is presented here that
dramatically limits circular boundary effects while preserving many of
the computational advantages of canonical frequency domain correlation filters. Our approach
demonstrated superior empirical results for both object detection and real-time
tracking compared to current state of the arts.

 {
\bibliographystyle{ieee}
\bibliography{egbib}

\begin{thebibliography}{10}\itemsep=-1pt

\bibitem{FragTrack}
A.~Adam, E.~Rivlin, and I.~Shimshoni.
\newblock Robust fragments based tracking using the integral histogram.
\newblock In {\em CVPR}, 2006.

\bibitem{Babenko-2009}
B.~Babenko, M.~H. Yang, and S.~Belongie.
\newblock Visual tracking with online multiple instance learning.
\newblock In {\em CVPR}, 2009.

\bibitem{Babenko2011}
B.~Babenko, M.-H. Yang, and S.~Belongie.
\newblock Robust object tracking with online multiple instance learning.
\newblock {\em PAMI}, 33(8):1619--1632, 2011.

\bibitem{Bolme10}
D.~S. Bolme, J.~R. Beveridge, B.~A. Draper, and Y.~M. Lui.
\newblock Visual object tracking using adaptive correlation filters.
\newblock In {\em CVPR}, 2010.

\bibitem{Bolme09}
D.~S. Bolme, B.~A. Draper, and J.~R. Beveridge.
\newblock Average of synthetic exact filters.
\newblock In {\em CVPR}, 2009.

\bibitem{boyd_book_2010}
S.~Boyd.
\newblock {Distributed Optimization and Statistical Learning via the
  Alternating Direction Method of Multipliers}.
\newblock {\em Foundations and Trends in Machine Learning}, 3:1--122, 2010.

\bibitem{delbue_PAMI_2011}
A.~{Del Bue}, J.~Xavier, L.~Agapito, and M.~Paladini.
\newblock {Bilinear Modelling via Augmented Lagrange Multipliers (BALM).}
\newblock {\em PAMI}, 34(8):1--14, Dec. 2011.

\bibitem{Grabner2006}
H.~Grabner, M.~Grabner, and H.~Bischof.
\newblock Real-time tracking via on-line boosting.
\newblock In {\em BMVC}, 2006.

\bibitem{Grabner2008}
H.~Grabner, C.~Leistner, and H.~Bischof.
\newblock Semi-supervised on-line boosting for robust tracking.
\newblock In {\em ECCV}. Springer, 2008.

\bibitem{Hare2011}
S.~Hare, A.~Saffari, and P.~H. Torr.
\newblock Struck: Structured output tracking with kernels.
\newblock In {\em ICCV}, 2011.

\bibitem{Henriques12}
J.~F. Henriques, R.~Caseiro, P.~Martines, and J.~Batista.
\newblock Exploiting the circulant structure of tracking-by-detection with
  kernels.
\newblock In {\em ECCV}, 2012.

\bibitem{Hester80}
C.~F. Hester and D.~Casasent.
\newblock Multivariant technique for multiclass pattern recognition.
\newblock {\em Appl. Opt.}, 19(11):1758--1761, 1980.

\bibitem{Kumar86}
B.~V. K.~V. Kumar.
\newblock Minimum-variance synthetic discriminant functions.
\newblock {\em J. Opt. Soc. Am. A}, 3(10):1579--1584, 1986.

\bibitem{CPR_Book}
B.~V. K.~V. Kumar, A.~Mahalanobis, and R.~D. Juday.
\newblock {\em Correlation Pattern Recognition}.
\newblock Cambridge University Press, 2005.

\bibitem{Mahalanobis87}
A.~Mahalanobis, B.~V. K.~V. Kumar, and D.~Casasent.
\newblock Minimum average correlation energy filters.
\newblock {\em Appl. Opt.}, 26(17):3633--3640, 1987.

\bibitem{Oza}
N.~C. Oza.
\newblock {\em {Online Ensemble Learning}}.
\newblock PhD thesis, U. C. Berkley, 2001.

\bibitem{Refregier91}
P.~Refregier.
\newblock Optimal trade-off filters for noise robustness, sharpness of the
  correlation peak, and horner efficiency.
\newblock {\em Optics Letters}, 16:829--832, 1991.

\bibitem{Ross_IJCV2008}
D.~{Ross}, J.~{Lim}, R.~{Lin}, and M.~{Yang}.
\newblock Incremental learning for robust visual tracking.
\newblock {\em IJCV}, 77(1):125--141, 2008.

\bibitem{Savvides03}
M.~Savvides and B.~V. K.~V. Kumar.
\newblock Efficient design of advanced correlation filters for robust
  distortion-tolerant face recognition.
\newblock In {\em AVSS}, pages 45--52, 2003.

\bibitem{zeiler_CVPR_2010}
M.~Zeiler, D.~Krishnan, and G.~Taylor.
\newblock {Deconvolutional networks}.
\newblock {\em CVPR}, 2010.

\end{thebibliography}
}

\end{document}